%% file: alphaZeroOthello_ToG.R3-arxiv.tex
\pdfoutput=1

\documentclass[conference]{IEEEtran}
\ifCLASSINFOpdf
\else
\fi

\usepackage{graphicx}

\pdfoutput=1
\usepackage{floatrow}
\usepackage[draft]{changes}

\usepackage{graphicx}
\usepackage{bm}
\usepackage[utf8]{inputenc}		
\usepackage[english]{babel}		
\usepackage{xspace}
\usepackage{amsmath}					
\usepackage{algorithm}
\usepackage{multirow}					
\usepackage{colortbl}					
\usepackage{tabularx}
\usepackage{subfigure}
\usepackage[noend]{algpseudocode}	
\usepackage{stackrel}

\usepackage{tikz}
\usetikzlibrary{shapes}
\usetikzlibrary{arrows}				
\usetikzlibrary{positioning}	
\usetikzlibrary{calc}					
\usetikzlibrary{matrix} 	    

\newcommand{\latexOrPdflatex}[2]{\ifx\undefined\pdfoutput%
#1%
\else%
#2%
\fi}

\latexOrPdflatex{
\newcommand{\href}[2]{#2}
}{
\usepackage[colorlinks,pagebackref,citecolor=blue,bookmarks=true,pdfstartview=FitH,pdfstartpage=1,pdftitle={General Board Game Playing Framework},pdfauthor={Johannes Scheiermann and Wolfgang Konen}]{hyperref}     
}

\newcommand{\TCLbase}{\textit{TCL-base}\xspace}
\newcommand{\TCLwrap}{\textit{TCL-wrap}\xspace}
\newcommand{\bfTheta}{\mathbf{\Theta}}

\newcommand{\AfterstateLogic}[1]{{afterstate logic}\xspace}
\newcommand{\afterstate}[1]{{afterstate}\xspace}
\newcommand{\EligibilityMethod}[1]{{eligibility method}\xspace}
\newcommand{\NPlayerGames}[1]{{$N$-player games}\xspace}
\newcommand{\FinalAdaptStep}[1]{{final adaptation step}\xspace}
\newcommand{\Jaskowski}[1]{Ja{\'s}kowski\xspace}

\newcommand{\REV}[1]{#1}                      
\newcommand{\REVB}[1]{#1}   									

\usepackage{changes}
\usepackage{fancyheadings}

\begin{document}
%
\title{
\REV{AlphaZero-Inspired Game Learning: Faster Training
by Using MCTS Only at Test Time}
}

\author{\IEEEauthorblockN{Johannes Scheiermann}
\IEEEauthorblockA{
RWTH Aachen University\\
Germany\\
Email: johannes.scheiermann@rwth-aachen.de
}
\and
\IEEEauthorblockN{Wolfgang Konen}
\IEEEauthorblockA{
TH Köln -- University of Applied Sciences\\
Germany\\
Email: wolfgang.konen@th-koeln.de
}
}


%


\maketitle
\begin{abstract}
Recently, the seminal algorithms AlphaGo and AlphaZero have started a new era in game learning and deep reinforcement learning. While the achievements of AlphaGo and AlphaZero -- playing Go and other complex games at super human level -- are truly impressive, these architectures have the drawback that they 
require high computational resources. Many researchers
are looking for methods that are similar to AlphaZero, but have lower computational demands and are thus more easily reproducible. \\
In this paper, we pick an important element of AlphaZero -- the Monte Carlo Tree Search (MCTS) planning stage -- and combine it with 
\REV{temporal difference (TD) learning} agents. We wrap MCTS for the first time around \REV{TD} n-tuple networks \REVB{and we use this wrapping only at test time} to create versatile agents that keep at the same time the computational demands low. We apply this new architecture to several complex games (Othello, ConnectFour, Rubik's Cube) and show the advantages achieved with this AlphaZero-inspired MCTS wrapper. In particular, we present results that this 
agent is the first one trained on standard hardware (no GPU or TPU) to beat the very strong Othello program Edax up to and including level 7 (where most other \REV{learning-from-scratch} algorithms could only defeat Edax up to level 2).

%

\end{abstract}




%
\IEEEpeerreviewmaketitle


\setlength{\headheight}{30pt}
\thispagestyle{fancy}
\lhead{}\chead{\footnotesize \copyright\ 2022 IEEE. This article has been accepted for publication in IEEE Transactions on Games. This is the author's version which has not been fully edited and content may change prior to final publication. Citation information: DOI 10.1109/TG.2022.3206733.}
\renewcommand{\headrulewidth}{0pt}

\section{Introduction} \label{sec:introduction}

\subsection{Motivation}

In computer science, game learning and game playing  are  interesting test beds for strategic decision making done by computers. Games usually have large state spaces, and they often require complex pattern recognition and strategic planning capabilities to decide which move is the best in a certain situation. If an algorithm is able to learn a game (or, even better, a variety of different games) just by self-play, given no other knowledge than the game rules, it is likely to perform also well on other problems of strategic decision making. 

With their seminal papers on AlphaGo~\cite{silver2016AlphaGo}, AlphaGo Zero~\cite{silver2017AlphaGoZero} and Alpha\-Zero~\cite{silver2017AlpaZeroChess}, Silver et al. opened a new door in game learning by presenting self-learning algorithms for the game of Go (which was considered to be unattainable for computers prior to theses publications). 
All these algorithms were able to beat the human Go world champion Lee Sedol. 

However, the full algorithms in ~\cite{silver2016AlphaGo}--\cite{silver2017AlpaZeroChess}
require huge computational resources in order to learn how to play the game of Go at world-master level. It is the purpose of this work to investigate whether some of the important elements of AlphaZero can already reach decent advances in game learning with much smaller computational efforts. For this purpose, we study several games -- namely Othello, ConnectFour and Rubik's Cube -- that have a lower complexity than Go
yet are not easy to master for both humans and game learning algorithms. The goal is to deliver not only agents with average game playing strength 
but agents that learn from scratch and play almost as well as the strongest known algorithms\footnote{\REVB{e.~g. Edax for Othello or AB-DL for ConnectFour as will be further explained in Sec.~\ref{sec:results} }} for these games.
We will show that this can be achieved for Othello and ConnectFour and, 
to some extent, also for Rubik's Cube. 

In this work, we pick an element of AlphaZero (here: the MCTS planning stage) and combine it
with reinforcement learning (RL) agents. 
We wrap MCTS 
around TD($\lambda$) n-tuple networks~\REV{\cite{Bagh15,Lucas08,szubert2014temporal}}, but the same technique could be applied to all types
of RL agents. 


The main contributions of this paper are as follows: (i) it shows for the first time -- to the best of our knowledge -- \REV{ a coupling between trainable TD($\lambda$) n-tuple networks} and MCTS planning; 
(ii) an AlphaZero-inspired solution, but with largely reduced computational requirements;  \REV{(iii) a comparison between MCTS inside and outside of self-play training;} (iv) very good results on Othello, ConnectFour and 2x2x2 Rubik's Cube. \REV{Specifically, we are able  to defeat the strong Othello program Edax at level~7 with an agent trained from scratch in less than 2 hours on a standard CPU.}

The rest of this paper is organized as follows: Sec.~\ref{sec:algoMeth} details the algorithmic building blocks and methods of our approach. Sec.~\ref{sec:experiment} describes the experimental setup, the games and the evaluation methods.
Sec.~\ref{sec:results} presents the results on the three games: quality achieved, interpretation, computation times. Sec.~\ref{sec:discuss} shows related work and discusses our results in comparison with other research. 
Sec.~\ref{sec:conclusion} concludes.

\section{Algorithms and Methods}
\label{sec:algoMeth}
The algorithm presented in this paper is implemented in the General Board Game (GBG) learning and playing framework ~\cite{Konen2019b,Konen22a},
which was developed for education and research in AI. GBG allows applying the new algorithm easily to a variety of games. GBG is
open source 
and available on  GitHub\footnote{\url{https://github.com/WolfgangKonen/GBG}}.

\subsection{Algorithm Overview}
\label{sec:newLogic}

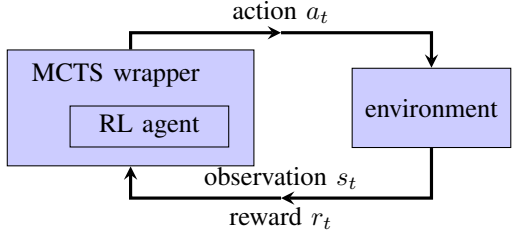
\begin{figure}[ht]%
\input{rl-with-mcts}
\caption{
Reinforcement learning with MCTS wrapper: The RL agent with MCTS wrapper observes a certain state $s_t$ and reward $r_t$  from the game environment and predicts the next action $a_t$.
}
\label{fig:RL-diagram}%
\end{figure}

The most important task of a game-playing agent is, given an observation or game state $s_t$ at time $t$, to propose a good next action $a_t$ from the set of actions available in $s_t$ (Fig.~\ref{fig:RL-diagram}). 
TD-learning uses the value function $V(s_{t})$, which is the expected sum of future rewards when being in state $s_{t}$. 

It is the task of the agent to learn the value function $V$ 
from experience (by interacting with the environment). In order to do so, it usually performs multiple self-play training episodes until a certain training budget is exhausted or a certain game-playing strength is reached. 

Our base RL algorithm TD-FARL is described in detail in \cite{konen2020reinforcement,konen2021final} and is partly inspired by Jaskowski et al.~\cite{jaskowski2018mastering}, van der Ree et al.~\cite{Ree2013reinforcement} and partly by our own experience with RL-n-tuple training. The key elements of the new RL-logic -- as opposed to our previous RL algorithms~\cite{Bagh15,Kone15c} -- are n-tuple systems, temporal coherence learning (TCL)~\cite{Beal99} and final adaptation RL (FARL)~\cite{konen2020reinforcement,konen2021final}. \REV{All these key elements will be briefly described in Sec.~\ref{sec:ntuples}, \ref{sec:td_farl} and \ref{sec:tcl}.} 

Despite being successful on a variety of games \cite{konen2020reinforcement,konen2021final}, this base algorithm shares one disadvantage with other deep learning algorithms that are only value-based: they base their decision on the value of the current state-action pairs. They have no planning component, no what-if scenarios to think about further consequences, like possible counter-actions of the other player(s), further own actions and so on. 

This is where AlphaZero's MCTS-trick
comes into play: Silver et al.~\cite{silver2016AlphaGo,silver2017AlphaGoZero} combine a deep learning RL agent with an MCTS wrapper (Fig.~\ref{fig:RL-diagram}) to introduce such a planning component. They do this throughout the whole training procedure,
which is better for the overall performance but 
also very computationally demanding. In this work, we take a simpler approach: we first train our RL agent, a TD n-tuple network, and then use the MCTS wrapping only at \REVB{test time (i.~e. during game play).} 
This usage of MCTS adds a form of planning at \REVB{test} time.

\subsection{MCTS Wrapper}
\label{sec:mctswrapper}


\REV{Our MCTS wrapper is based on}
 the UCT variant of MCTS, where the probability of predicting an optimal move converges to 100\% in the limit of an infinite number of iterations~\cite{kocsis2006mctsUCT}. 
If we limit the iterations to a fixed number, we only approach optimality but have a fixed runtime.

Therefore, with an MCTS,
promising results can 
be expected under reasonable computational requirements, given the number of MCTS iterations is correctly balanced.


The \REV{iterations} of MCTS usually consist of four consecutive steps: selection, expansion, simulation, and backpropagation~\cite{browne2012MCTS}. The following child selection policy, which is the one used by Silver et al.~\cite{silver2017AlphaGoZero} in AlphaGo Zero, is also the one we implemented in our MCTS wrapper for the same purpose:

\begin{eqnarray}
  a_t &=& \arg\max_{a \in A(s_t)}\left(\frac{\REV{W}(s_t,a)}{N(s_t,a)}+U(s_t,a)\right)
  \label{eq:UCB} \\
	U(s,a) &=& c_{puct}P(s,a)\frac{\sqrt{\varepsilon+\sum_{b \in A(s)}{N(s,b)}}}{1+N(s,a)}
	\label{eq:UCB2}
\end{eqnarray}

Here, $\REV{W}(s,a)$ is the accumulator for all backpropagated values (as detailed in Algorithm \ref{alg:mcts_iter} below) that arrive along with branch $a$ of node \textsc{r} that carries state $s$. Likewise, $N(s,a)$ is the visit counter and $P(s,a)$ the prior probability. $A(s)$ is the set of actions available in state $s$. 
$\varepsilon$ is a small positive constant for the special case $\sum_b{N(s,b)}=0$: It guarantees that in this special case the maximum of $U(s,a)$ is given by the maximum of $P(s,a)$. The prior probabilities $P(s,a)$ are obtained by sending the \REV{RL agent's values of all available state-action pairs $(s,a)$ with $a \in A(s)$} through a softmax function (see Sec.~\ref{sec:ntuples}).\footnote{Note that the prior probabilities and the MCTS iteration are only needed at test time, so that we -- different to AlphaZero -- do not need MCTS during self-play training.} 

According to Silver et al.~\cite{silver2017AlphaGoZero}, the above child selection policy is a variant of the PUCB ("Predictor + UCB") algorithm presented by Rosin~\cite{rosin2011multi-armed}.
Furthermore, the latter is a modification of the bandit algorithm UCB1, extending it with the behavior to also consider the recommendations of a predictor. UCB1 is also the basis of the previously mentioned algorithm UCT (UCB applied to trees) by Kocsis and Szepesvári~\cite{kocsis2006mctsUCT}.

\begin{algorithm}[tbp]
\caption{\textsc{MctsIteration}: This recursive algorithm is applicable to 1- or 2-player games. It performs a single iteration of a Monte Carlo tree search, starting from root node \textsc{r} carrying state $s$.}
\label{alg:mcts_iter}
   \begin{algorithmic}[1]
			\Function{MctsIteration}{\textsc{Node r}}
			    \State $\kappa=(-1)^{N-1}$  \Comment{$N$: number of players}

                \If{\textsc{IsGameOver}$(s)$}
                    \State\Return $\kappa*\textsc{FinalGameScore}(s)$
                \EndIf
                
                \If{$\textsc{r}.\textsc{expanded}=\textsc{false}$}
                    \State $(V,\mathbf{p}) \leftarrow f(s)$   \Comment{$f$: approximator network}
                    \State $P(s,\cdot) \leftarrow \mathbf{p}$  \Comment{prior probabilities given by $f$}
                    \State $\textsc{r}.\textsc{expanded} \leftarrow \textsc{true}$
                    \State\Return $\kappa*V$
                \EndIf 
                \\
                \State $(a,\textsc{c}) \leftarrow \textsc{SelectChild}(\textsc{r})$ \Comment{use Eq.~\eqref{eq:UCB} to select} 
								\State  																														\Comment{action $a$ and child \textsc{ c}}
                \State $V_{child} \leftarrow \textsc{MctsIteration}(\textsc{c})$
                \State $\REV{W}(s,a) \leftarrow \REV{W}(s,a) + V_{child} $
                \State $N(s,a)  \leftarrow N(s,a) + 1$
                \State
                \State\Return $\kappa*V_{child}$
			\EndFunction
	\end{algorithmic}
\end{algorithm}

Our implementation of an MCTS iteration is illustrated in Algorithm \ref{alg:mcts_iter}.
It performs a single MCTS iteration for a given node. The numerical return value approximates how valuable it is to choose an action that leads to this node. Since this assessment corresponds to the view of the previous player, the algorithm negates  the returned values ($\kappa=-1$) in the case of 2-player games.

If the node represents a game-over state, then the consequence of choosing this node is known and does not need to be approximated. In this case, the final game score is the value to propagate back.

Reaching a non-expanded node is also a termination condition. In this case, the approximator function $f$ (usually the wrapped RL agent of Fig.~\ref{fig:RL-diagram}) approximates the value $V$ of the corresponding node together with its action  probabilities $\mathbf{p}$ (line 6). 
Afterward, the node is marked as expanded, and its approximated value is propagated back. 

\textsc{SelectChild} is used to select a child node based on the PUCB variant of Eq.~\eqref{eq:UCB} 
if no previous termination condition occurred. To determine the selected child node's value, it serves as input to another recursive call of the \textsc{MctsIteration} algorithm. On return from the recursive call, the returned value $V_{child}$ is \REVB{added} to $\REV{W}(s,a)$ (line 14), and the visit count $N(s,a)$ is incremented.

Our MCTS implementation first performs a certain number of iterations starting from the node corresponding to the current game state in a concrete match. Then it decides on the action that leads to the most frequently visited child node.

Furthermore, our tree search \REV{adopts another element from AlphaGo} which reuses the previously built search tree whenever possible, i.~e., when a node corresponding to the current game state is already present in the search tree of the previous move. This optimization avoids performing superfluous MCTS iterations, \REV{which only recalculate already known facts again.}  


\begin{table*}[htb]
    \caption{
    N-tuple systems used in this work. Parameters $n$, $m$ and $k$ are explained in the main text. The number of weights is $4^7\cdot 200$ (Othello) and $4^8\cdot 140$ (ConnectFour). For 2x2x2 Rubik's Cube, each 7-tuple has either 3 or 7 positional values, depending on the cell location. Thus, the number of weights depends on the cell location. Since not every weight represents a reachable position, the number of active weights is smaller, as given by the percentage in the last column. } 
    \label{tab:ntuple-systems}
    \centering
    \begin{tabular}{|l|c|c|c|c|c|} \hline\hline
    game & length $n$ & position $m$ &  $k$ & weights & percent active \\ \hline\hline
    Othello & 7 & 4 & $2\cdot 100$& 3,276,800 &51\%\\
ConnectFour & 8 & 4 & $2\cdot 70$ & 9,175,040 & 8\%\\
2x2x2 Rubik's%
    & 7 & \{3, 7\} & 60 & 3,720,780 & 31\%\\
3x3x3 Rubik's 
    & 7 & \{2, 3, 8, 12\} &120 &46,563,392 & 22\%\\
 \hline\hline
    \end{tabular}
\end{table*}

\begin{figure}%
	\includegraphics[width=0.67\columnwidth]{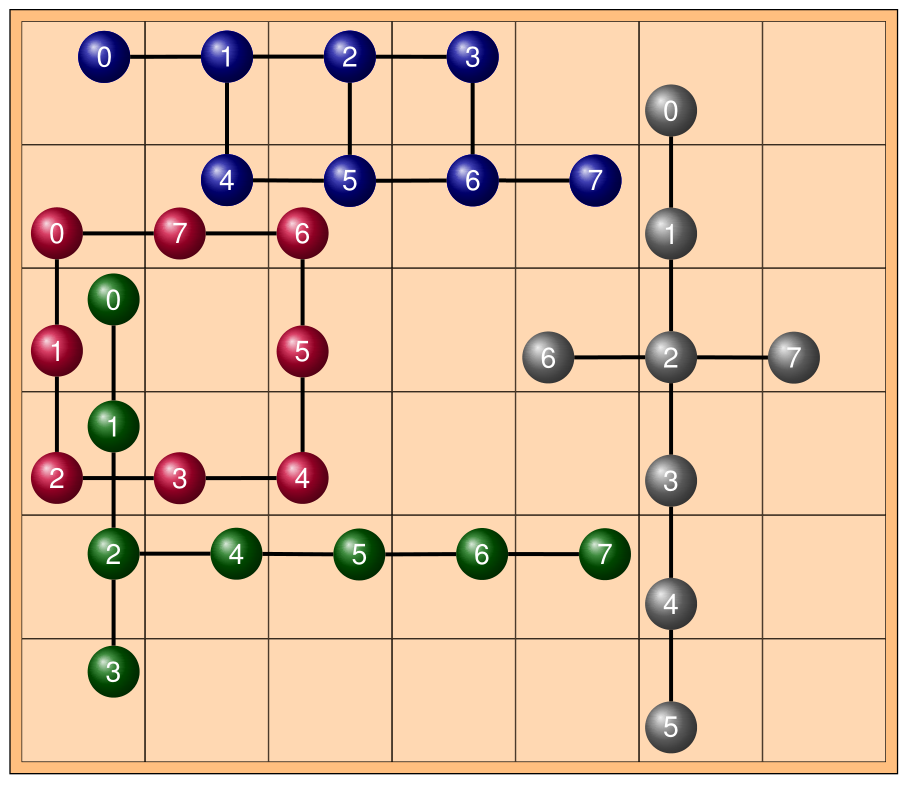}		
\caption{\REVB{Example n-tuples: We show 4 random-walk 8-tuples on a 6x7 ConnectFour board. The tuples are selected manually to show that not only snake-like shapes are possible, but also bifurcations or cross shapes. Tuples may or may not be symmetric.}}
\label{fig:ntuple01}%
\end{figure}

\subsection{N-Tuple Systems}
\label{sec:ntuples}
N-tuple systems coupled with TD were first applied to game learning by Lucas in 2008~\cite{Lucas08}, although n-tuples were already introduced in 1959 for character recognition purposes~\cite{bledsoe1959pattern}. The remarkable success of n-tuples in learning to play Othello~\cite{Lucas08} motivated other authors to benefit from this approach for a number of other games. The main goal of n-tuple systems is to map a highly non-linear function in a low dimensional space to a high dimensional space where it is easier to separate `good' and `bad' regions. This can be compared to the kernel trick of support-vector machines. An n-tuple is defined as a sequence of $n$ cells of the board. Each cell can have $m$ positional values representing the possible states of that cell. Therefore, every n-tuple will have a (possibly large) look-up table indexed in 
form of an $n$-digit number in base $m$. Each entry corresponds to a feature and carries a trainable weight. An n-tuple system is a system consisting of $k$ n-tuples. Tab.~\ref{tab:ntuple-systems} shows the n-tuple systems that we use in this work. Each time a new agent is constructed, all n-tuples are formed by \textit{random walk}. That is, all cells are placed randomly with the constraint that each cell must be adjacent\footnote{The form of adjacency (e.~g. 4- or 8-point neighborhood) is user-defined.} to at least one other cell in the n-tuple.
\REV{An example n-tuple system is shown in Fig.~\ref{fig:ntuple01}. 
}


\REV{
Let $\bfTheta$ be the vector of all weights $\theta_i$ of the n-tuple system. The length of this vector may be large, as the fifth column in Tab.~\ref{tab:ntuple-systems} shows. If all n-tuples have the same $n$ and $m$, $\bfTheta$ has length $m^n k$.
Let $\mathbf{\Phi}(s)$ be a binary vector of the same length representing the feature occurences in state $s$. The value function of the n-tuple network given state $s$ is 
\begin{equation}
		V(s) = \sigma \left( \mathbf{\Phi}(s)\cdot \bfTheta\right)
\label{eq:valueNtuple}
\end{equation}
with transfer function $\sigma$ which may be a sigmoidal function or simply the identity function.
} 

\REV{An agent using this n-tuple system derives a policy from the value function in Eq.~\eqref{eq:valueNtuple} as follows: Given state $s$ and the set $A(s)$ of available actions in state $s$, it applies with a forward model $f$
every action $a \in A(s)$ to state $s$, yielding the next state $s' = f(s,a)$. Then it selects the action that maximizes $V(s')$.}

\REV{The prior probabilities $P(s,a)$ in Eq.~\eqref{eq:UCB2} are calculated similarly: Given state $s$ and actions $a \in A(s)$, we compute with Eq.~\eqref{eq:valueNtuple} all values $V(s')$ of reachable states $s' = f(s,a)$ and send them through a softmax function to yield $P(s,a)$.
}

\subsection{\REV{TD Learning and FARL}}
\label{sec:td_farl}
\REV{The goal of n-tuple agent training is to learn a value function~$V$ that generates a good policy, i.~e. a policy that selects in almost all cases the best action. 
In our work, we use the TD learning algorithm TD-FARL~\cite{konen2020reinforcement,konen2021final} for training, which is briefly described in the following.
} 

\REV{Let $s'[p]$ be the actual state generated by acting player $p$ and let $s[p]$ be the previous state that was generated by this acting player. TD(0) learning \cite{Ree2013reinforcement,SuttBart98} adapts the value function with model parameters $\bfTheta$ through
\begin{equation}
		\bfTheta_{t+1} = \bfTheta_{t} + \alpha\delta\mathbf{\nabla_{\bfTheta}} V(s[p])
\label{eq:theta}
\end{equation}
Here, $\alpha$ is the learning rate and $V$ is in our case the n-tuple value function of  Eq.~\eqref{eq:valueNtuple}. $\delta$ is the usual TD error \cite{SuttBart98} after player $p$ has acted and generated $s'[p]$:
\begin{equation}
		\delta = r[p]+\gamma V(s'[p]) - V(s[p])
\label{eq:TDdelta}
\end{equation}
where the sum of the first two terms (reward $r[p]$ for player $p$ and the discounted value $\gamma V(s'[p])$) is the desirable target for $V(s[p])$.
}

\REV{The extra element of FARL is to add a final adaptation step at the end of each episode: All players $p'$ \textit{different} from the last player take \textit{their} reward $r[p']$ and adapt the value function for state $s[p']$ according to Eq.~\eqref{eq:theta}, but with error signal
\begin{equation}
		\delta = r[p'] - V(s[p'])  \quad \forall p' \neq p
\label{eq:TDdeltaFARL}
\end{equation}
This FARL step was found to be crucial for reaching training success in $N$-player games with arbitrary $N$~\cite{konen2020reinforcement}. 
}

\REV{TD-FARL performs its training purely by $\epsilon$-greedy self-play: The agent plays against itself and makes random moves with probability $\epsilon$ in order to explore. No external knowledge or databases are used. More details on TD-FARL for $N$-player games, including the extension to TD($\lambda$) with arbitrary $\lambda \in [0,1]$, based on the eligibility mechanism for n-tuple systems~\cite{jaskowski2018mastering}, are described in our previous work \cite{konen2021final}.
}
\vspace{0.25cm}

\subsection{Temporal Coherence Learning (TCL)} 
\label{sec:tcl}
The TCL algorithm developed by Beal and Smith~\cite{Beal99} is an extension of TD learning. It replaces the global learning rate $\alpha$ with the weight-individual product $\alpha\alpha_i$ for every weight $\REV{\theta}_i$. Here, the adjustable learning rate $\alpha_i$ is a free parameter set by a pretty simple procedure: For each weight $\REV{\theta}_i$, two counters $N_i$ and $A_i$ accumulate the sum of weight changes and the sum of absolute weight changes. If all weight changes have the same sign, then $\alpha_i=|N_i|/A_i=1$, and the learning rate stays at its upper bound. If weight changes have alternating signs, then the global learning rate is probably too large. In this case, $\alpha_i=|N_i|/A_i \rightarrow 0$ for $t \rightarrow \infty$, and the effective learning rate will be largely reduced for this weight. 

\REV{In our previous work~\cite{Bagh15} we extended TCL to $\alpha_i=g(|N_i|/A_i)$ where $g$ is a transfer function being either the identity function (standard TCL) or an exponential function $g(x)=e^{\beta(x-1)}$.
It was shown in~\cite{Bagh15}} that TCL leads to faster learning and higher win rates for the game ConnectFour.

\section{Experimental Setup}
\label{sec:experiment}

\subsection{The Games}
\label{sec:games}

\begin{figure}[tbp]
\centerline{
    \begin{tabular}{cc}
    \includegraphics[width=0.45\columnwidth]{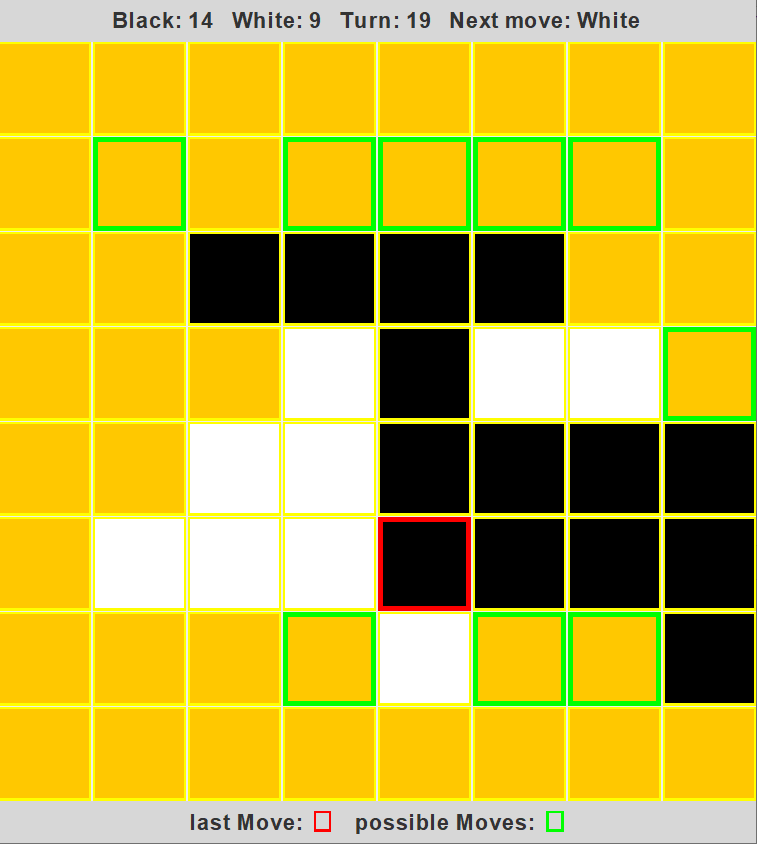}
             &  \hspace{0.05\columnwidth}
    \includegraphics[width=0.45\columnwidth,height=0.41\columnwidth]{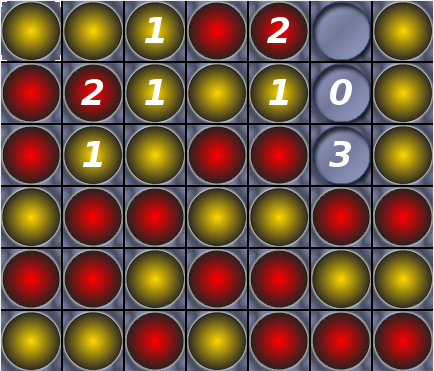}
         \\
 (a)     &    \hspace{0.05\columnwidth} (b) 
    \end{tabular}
} 
\caption{(a) Othello game state. The black cell with a red outline marks the last move of Black. It is White's turn to choose one of the available actions marked by cells with a green border. These actions capture one or more black pieces, which are then flipped to white. (b) ConnectFour game state. It is Red's turn, and \REV{they have} to place \REV{their} piece in the only free column. Subsequently, Yellow wins by reaching \textit{Four in a Row}. Numbers show cell coding: 1 and 2 for players' pieces, 3: empty and reachable, 0: empty, but not reachable (in next move).
}
\label{fig:gamesOthelloC4}
\end{figure}

\subsubsection{Othello}
\label{sec:gameOthello}
(Reversi) is a well-known board game with quite simple rules yet requiring complex strategies to play strongly. Fig.~\ref{fig:gamesOthelloC4}(a) shows a typical game position. The regular 8x8 Othello has $10^{28}$ states and an average branching factor of 10. It is an unsolved game (no perfect winning strategy is known).

\REV{In our Othello experiments, we compete against Edax~\cite{delorme2019}, a strong Othello playing program. We use Edax in the same way as~Norelli et al.~\cite{norelli2022olivaw} with standard settings (alpha-beta search, tabular value functions, no opening books) and vary only the Edax level (search depth).} 
\subsubsection{ConnectFour}
\label{sec:gameC4}
 (Four in a Row) is another board game with quite simple rules. Fig.~\ref{fig:gamesOthelloC4}(b) shows a typical end game position. The regular 6x7 ConnectFour has $10^{12}$ states and a branching factor $\leq 7$. It is a solved game: The $1^{st}$ player wins if playing perfectly.

\REV{In our ConnectFour experiments, we compete against the strong alpha-beta search agents AB and AB-DL~\cite{Thil14},  further described in Sec.~\ref{sec:result-C4}. AB and AB-DL work with a pre-computed opening book and do not need any evaluation functions, search depth or time budget parameters.} 

\subsubsection{Rubik's Cube}
\label{sec:gameCube}
 is a famous puzzle  \REV{of a cube consisting of smaller 'cubies'} where the goal is to move an arbitrary scrambled cube 
into the solved position through a sequence of twists. \REV{In the solved position, each cube face consists of 9 (3x3x3 cube) or 4 (2x2x2 cube) cubie faces with the \textit{same}  color.} The regular 3x3x3 cube has $4.3\cdot 10^{19}$ states and a branching factor of $18$. 
The 2x2x2 cube has $3.6\cdot 10^{6}$ states and a branching factor of 9.

\subsection{Common Settings}

We use for all our experiments the same RL agent based on n-tuple systems and TCL. Only its hyperparameters are tuned to the specific game, as shown below. We refer to this agent as \textbf{\TCLbase} whenever it alone is used  for game playing. If we wrap this agent by an MCTS wrapper with a given number of iterations, then we refer to this as \textbf{\TCLwrap}.

The hyperparameters for each game were found by manual fine-tuning. \REV{
For reasons of space we give the exact explanation and the setting of all parameters
as supplementary material in Appendix G of~\cite{Konen22a}. 
A short version of these settings is:} The chosen n-tuple configurations are given in Tab.~\ref{tab:ntuple-systems}, and the main parameters are:
\begin{itemize}
\item \textbf{Othello}: learning rate $\alpha=0.2$, TCL activated, 
eligibility trace factor $\lambda=0.5$, exploration rate $\epsilon=0.2 \rightarrow 0.1$, 250,000 training episodes.
\item \textbf{ConnectFour}: learning rate $\alpha=3.7$, TCL activated, 
eligibility trace factor $\lambda=0.0$, exploration rate $\epsilon=0.1 \rightarrow 0.0$, 6,000,000 training episodes.
\item \textbf{Rubik's Cube}: learning rate $\alpha=0.25$, TCL activated,  
eligibility trace factor $\lambda=0.0$, exploration rate $\epsilon=0.0$, 3,000,000 training episodes.
\end{itemize}


\begin{figure}[htb]%
\centerline{
\includegraphics[width=1.0\columnwidth]{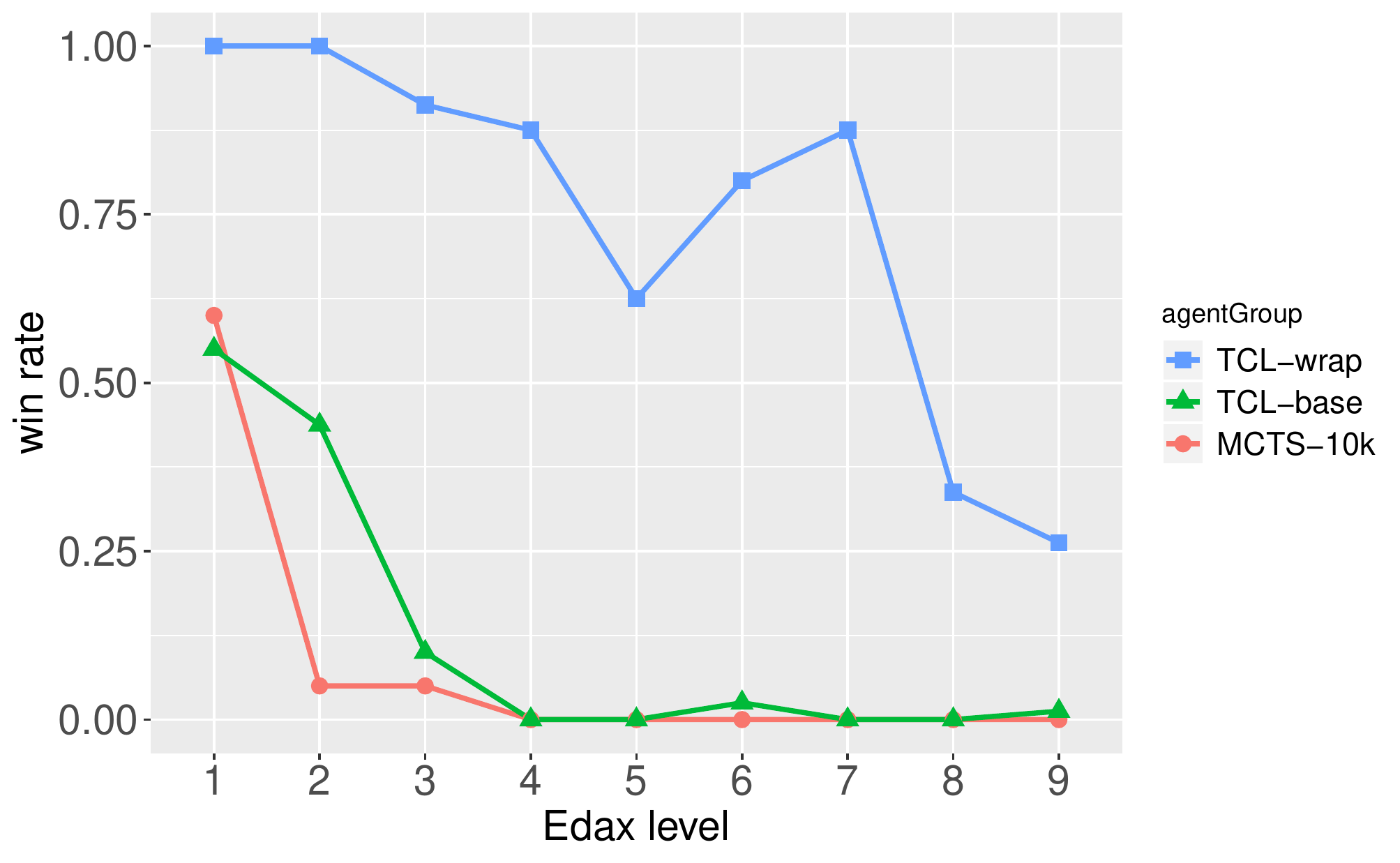}}
\caption{
Different Othello agents playing against Edax. \textit{\TCLwrap}: TCL coupled with MCTS wrapper (10,000 iterations); \textit{\TCLbase}: TCL alone; \textit{MCTS-10k}: MCTS alone with 10,000 iterations. 
\REV{Shown are the percentages of won runs from} (a) 20 TCL agents trained with different random n-tuples in the TCL cases and (b) 20 MCTS agents with different seeds in the MCTS case. Each agent plays in both roles (1${}^{st}$ and 2${}^{nd}$ player), \REV{yielding 40 competition runs in total.}
}
\label{fig:MCTSWrap-TCL}%
\end{figure}

\section{Results}
\label{sec:results}

\subsection{Othello}
It is not too difficult for game learning algorithms to reach a medium playing strength in Othello, i.~e. a strength where simple heuristic players are beaten~\cite{Lucas08,wang2020warm}. But it is very difficult to beat the very strong Othello playing program  Edax~\cite{delorme2019}. Edax has a configurable  playing strength (level, depth) between 0 and 60. 

We compare our agents with Edax at different levels. Since all agents (Edax, \TCLbase and \TCLwrap) are  deterministic move predictors, repeated evaluation runs with the same pair of agents always yield the same results and cannot be used to collect statistics. We use the following procedure to get statistically sound results: We draw 20 different random n-tuple configurations (\textit{random walk}, see Sec.~\ref{sec:ntuples})
and train for each configuration a separate \TCLbase agent. \REV{All TCL agents compete in both roles against Edax, yielding 40 competition runs.} 

Fig.~\ref{fig:MCTSWrap-TCL} shows the \REV{resulting win rates (win count divided by 40 runs)}: Both MCTS and \TCLbase cannot defeat Edax at level 2 and above (their win rates are lower than 50\% from level 2 on). The situation changes dramatically as soon as we wrap \TCLbase by MCTS: \TCLwrap defeats Edax up to level 7 and has win rates above 25\% for levels 8 and 9. \REV{The non-monotonuous trend of \TCLwrap at Edax level 5 and 6 is surprising and not fully understood: It could be statistical fluctuations or it could be that Edax plays a little weaker at level 7 than at level 5 or 6.}

\begin{figure}[htb]
\centerline{
    \begin{tabular}{cc}
    \includegraphics[width=0.45\columnwidth]{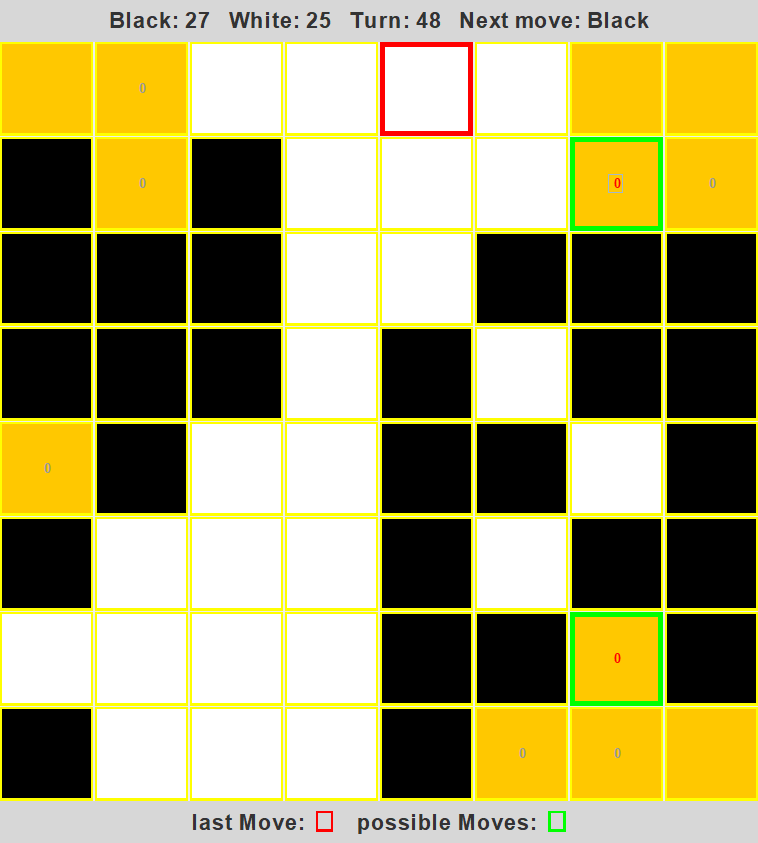}
             &  \hspace{0.05\columnwidth}
    \includegraphics[width=0.45\columnwidth]{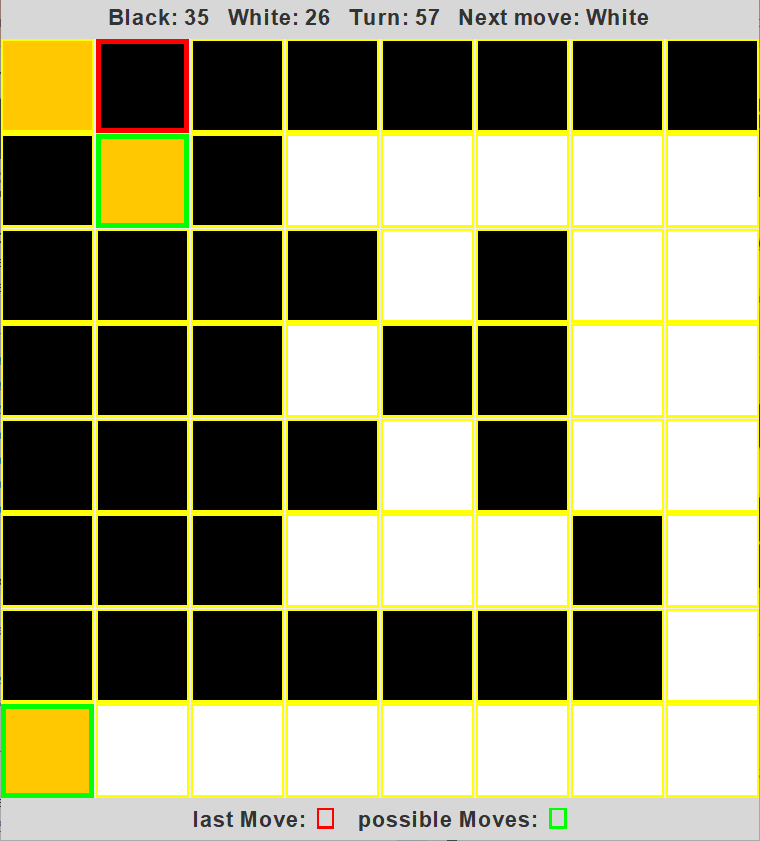}
         \\
 (a)     &    \hspace{0.05\columnwidth} (b) 
    \end{tabular}
} 
\caption{Tactics of Edax in Othello: (a) Move 48 in a game \TCLbase (Black) vs. Edax level 7 (White): It is Black's turn, and Edax forces Black into disadvantageous moves that allow White to capture the corners. \TCLwrap will avoid such disadvantageous positions. (b) Move 57 in a game \TCLwrap (Black) vs. Edax level 8 (White): Now it is White's turn, and although Black has the current majority of pieces, White will eventually win because Black has to pass and White moves \textit{three times in a row}.
}
\label{fig:Edax-tactics}
\end{figure}

\textit{Interpretation}: What are the reasons for opponents to win or lose in Othello against Edax? -- To investigate this, we analyze specific Othello episodes: When Edax plays at level 7, it has advanced tactics that narrow the range of possible actions for the opponent (Fig.~\ref{fig:Edax-tactics}(a)): If Edax ($2^{nd}$) plays against opponent \TCLbase ($1^{st}$), Edax forces \TCLbase towards the end of the episode to play disadvantageous moves. If we now replace the opponent ($1^{st}$) with \TCLwrap, it avoids these traps: The planning stage of \TCLwrap helps to foresee the disadvantageous positions when they are some moves ahead; now \TCLwrap finds other moves to avoid them and is thus not forced into the disadvantageous positions. 

At level 8 or higher, Edax shows \REV{sometimes} another tactic: It \REV{may play} in such a way that the last 2-4 moves are pass moves for the opponent (Fig.~\ref{fig:Edax-tactics}(b)): Since the opponent has no available action at its disposal, it is forced to pass the move right to Edax again. During the very last moves of an episode, Edax \REV{may} gain the majority of pieces. Currently, \TCLwrap is not able to avoid these pass situations, at least not in the majority of the episodes played. 

\begin{table*}[htb]
    \caption{
    Results of a ConnectFour tournament with 4 agents. Shown is the W/T/L  (win/tie/loss) count of the $1^{st}$ (row) player when playing 100 episodes against the $2^{nd}$ (column) player. Agents are ranked
    by their overall rate of games won (last column). The colored and bold numbers show remarkable improvements of \TCLwrap over \TCLbase.} 
    \label{tab:C4-4agent-tourn}
    \centering
    \begin{footnotesize}
    \begin{tabular}{|l|l|c|c|c|c|c|} \hline\hline
     \multicolumn{2}{|c|}{\multirow{2}{*}{W/T/L}}  & \multicolumn{4}{|c|}{$2^{nd}$ player}     
                 & \multirow{2}{*}{won games rate}  \\ \cline{3-6}
     \multicolumn{2}{|c|}{} 
          & \TCLwrap & \ AB-DL\ \ & \TCLbase &  \ MCTS\ \ \ & \\ \hline \hline
    \multirow{4}{*}{$1^{st}$} 
    & \TCLwrap &  & 99/0/\textbf{\color{blue}1}  & 100/0/0  & 100/0/0  & 66.3\% \\ \cline{2-7}
    & AB-DL & 100/0/0 &   & 100/0/0  & 100/0/0    & 64.9\% \\ \cline{2-7}
    & \TCLbase & 100/0/0  & 91/2/\textbf{\color{blue}7} &   & 100/0/0  & 49.0\% \\ \cline{2-7}
    & MCTS &  \textbf{\textcolor[rgb]{0,0.7,0}{1}}/2/97  & 17/3/80  &  \textbf{\textcolor[rgb]{0,0.7,0}{97}}/2/1  &      & 19.8\% \\ \hline\hline
    \end{tabular}
    \end{footnotesize}
\end{table*}

\subsection{ConnectFour} 
\label{sec:result-C4}

ConnectFour is a non-trivial game that is not easy to master for humans. However, its medium-size complexity allows for very strong tree-based solutions when combined with a pre-computed opening book. These near-perfect agents are termed AB and AB-DL since they are based on alpha-beta search (AB) that extends the Minimax algorithm by efficiently pruning the search tree. Thill et al.~\cite{Thil14} were able to implement alpha-beta search for  ConnectFour in such a way that it plays near-perfect: It wins all games as $1^{st}$ player and wins very often as $2^{nd}$ player when the $1^{st}$ player makes a wrong move.
AB and AB-DL  differ in the way they react to losing states: While AB just takes a random move, AB-DL searches for the move, which postpones the loss as far (as distant) as possible (DL = distant losses). It is tougher to win against AB-DL since it will request more correct moves from the opponent and will very often punish wrong moves.

We perform a tournament with the following 4 agents:
\begin{itemize}
    \item \textbf{\TCLwrap}: MCTSWrapper[\textit{TCL-base}]
    (iter=1,000, $c_{PUCT}$=1.0, unlimited depth), 
    \item \textbf{\TCLbase}: TCL alone, 
    \item \textbf{AB-DL}: Alpha Beta with Distant Losses,
    \item \textbf{MCTS}: MCTS(UCT, random playouts, iter=10,000, treeDepth=40) 
\end{itemize}

\begin{figure}[htb]%
\centerline{
\includegraphics[width=1.0\columnwidth]{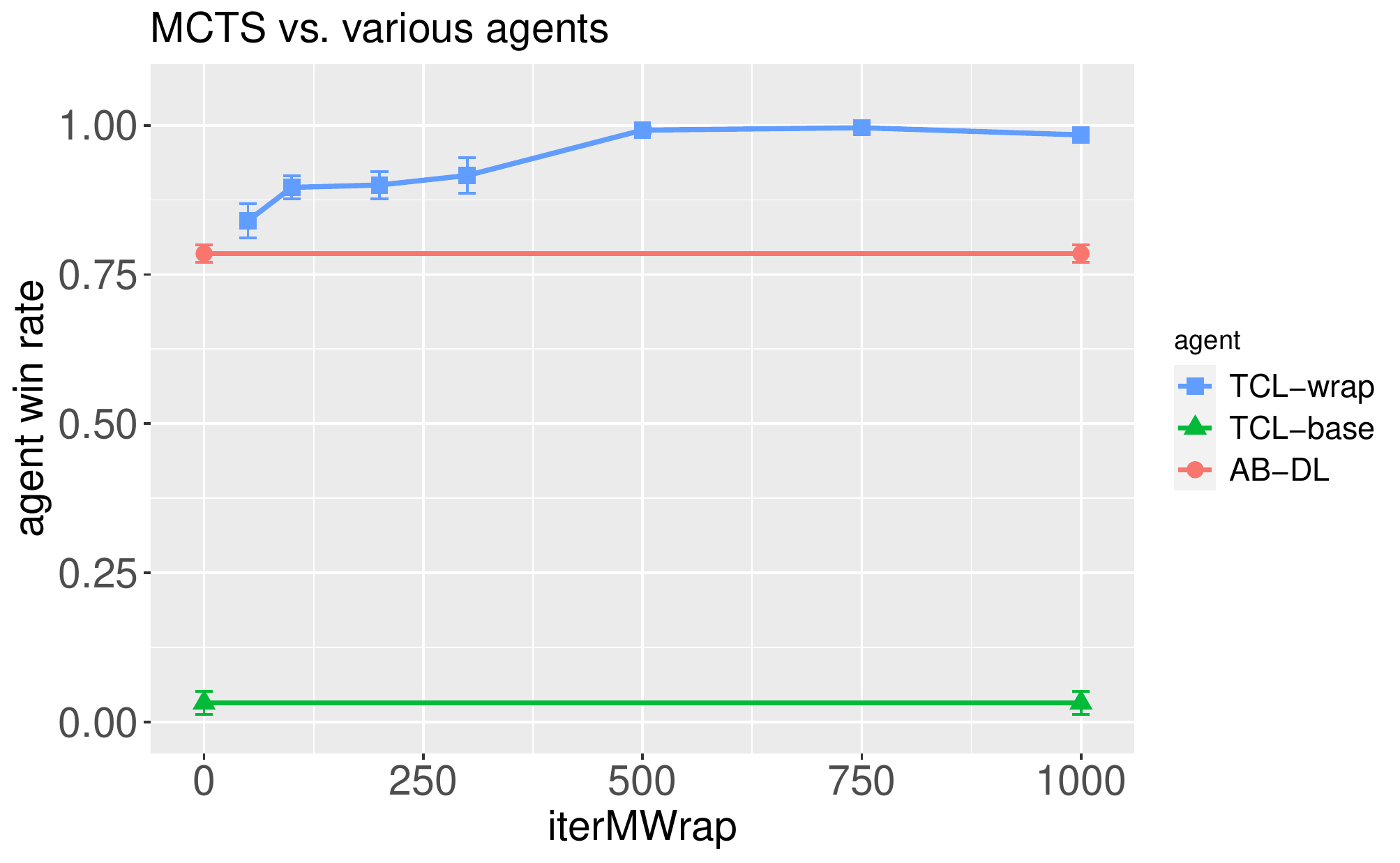}}
\caption{
The effect of MCTS wrapping on ConnectFour. Shown are the averages from 5 runs, where each run consists of 25 competition episodes MCTS ($1^{st}$) vs. \REVB{agent} ($2^{nd}$). The \REVB{agents} are a)
\textit{\TCLwrap}: TCL wrapped by MCTS wrapper with iterMWrap iterations; b) \textit{AB-DL}: Alpha-Beta agent with distant losses; c) \textit{\TCLbase}: TCL without MCTS wrapper. \REV{Error bars denote standard deviations.} }
\label{fig:C4-MCTS-vs-Opponent}%
\end{figure}

The results are shown in Tab.~\ref{tab:C4-4agent-tourn} and can be described as follows: \TCLwrap and AB-DL win nearly all their games when playing first (ConnectFour is a theoretical win for the $1^{st}$ player). \TCLbase ($1^{st}$) wins against AB-DL ($2^{nd}$) the majority of its games (91\%), but not all. If we enhance \TCLbase by MCTS wrapper, the win rate of \TCLwrap rises to fantastic 99\%, so it avoids 8/9 
of the former \TCLbase losses or ties. 

MCTS, as the weakest agent in the tournament, wins as $1^{st}$ player most of its games (97\%) against \TCLbase ($2^{nd}$), but it predominantly loses against \TCLwrap and AB-DL ($2^{nd}$). \TCLwrap as $2^{nd}$ player is in this respect significantly stronger than AB-DL (97\% vs. 80\% win rate, resp.), which leads for \TCLwrap to a higher total rate of 66.3\% won games as compared to AB-DL (64.9\%). Besides that, the total won games rate 66.3\% is a big jump forward when compared to the total won game rate 49\% of \TCLbase.

\textit{Interpretation}: MCTS plays differently, perhaps more surprising, than near-optimal agents. Since \TCLbase was trained on a near-optimal agent (itself), it has never seen the `surprising' moves of MCTS and will probably often react wrongly on these moves. Thus, \TCLbase loses most of its games when playing $2^{nd}$. If we now add with MCTS wrapper a planning component to \TCLbase, then \TCLwrap can find better responses to the `surprising' moves, and it can better exploit the occasional wrong moves of MCTS. As a consequence, it wins most of the episodes.

Fig.~\ref{fig:C4-MCTS-vs-Opponent} shows the results of MCTS wrapping in ConnectFour as a function of MCTS wrapper iterations. Even a small amount of iterations (50-100) already leads to a  \TCLwrap win rate of $> 80\%$. With 500 iterations or more, \TCLwrap achieves a win rate near 100\%. 

\REV{Additionally, we made an experiment comparing our agents AlphaBeta-DL and \TCLwrap with ConnectZero from Dawson \cite{dawson2020}: We performed 10 episodes with ConnectZero starting (which is a theoretical win), but found instead that AlphaBeta playing second won 80\% of the episodes and \TCLwrap playing second won all episodes.}

\begin{figure}[htb]%
\centerline{
    \begin{tabular}{c}
\includegraphics[width=1.0\columnwidth]{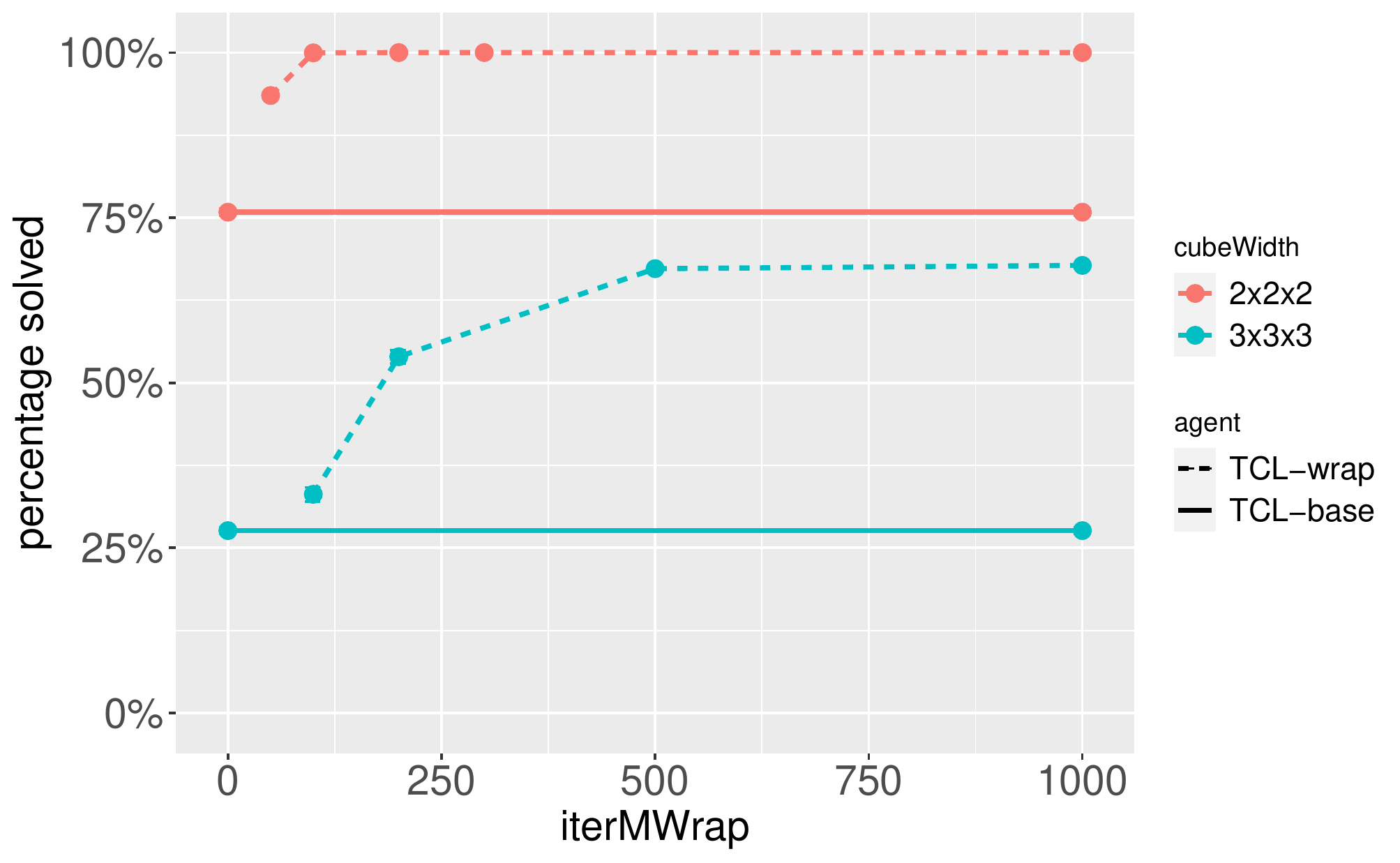}
        \\
				{\footnotesize (a)}
				\\
\includegraphics[width=1.0\columnwidth]{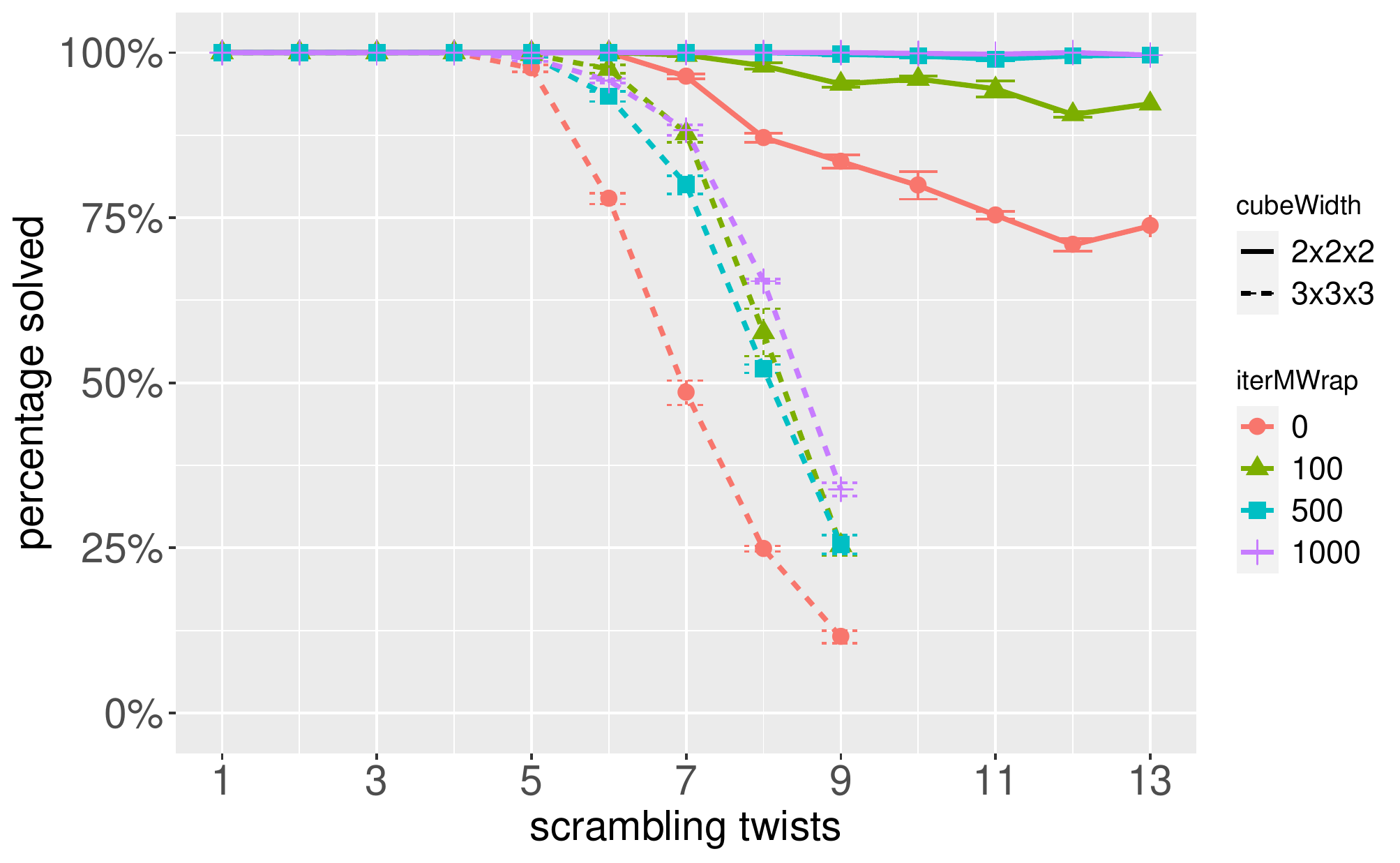}
         \\
				{\footnotesize (b)}
    \end{tabular}
} 
\caption{
The effect of MCTS wrapping on Rubik's Cube. Shown are the averages from 4 runs, where each run evaluates the ability of the agents to solve a large set of scrambled cubes (a) as a function of MCTS iterations: 2x2x2: 600 cubes scrambled with either 11, 12 or 13 twists; 3x3x3: 600 cubes scrambled with either 7, 8 or 9 twists; (b) as a function of scrambling twists 
\REV{where iterMWrap=0 corresponds to \TCLbase and iterMWrap$>$0 to different variants of \TCLwrap. We average over} 200 scrambled cubes for each twist number.  \REV{Error bars denote standard deviations.}
}
\label{fig:Rubiks-2x2-3x3}%
\end{figure}

\subsection{Rubik's Cube}

We investigate two variants of Rubik's Cube: 2x2x2 and 3x3x3. We trained TCL agents by presenting them cubes scrambled with up to $p_{max}$ twists where $p_{max}=13$ for 2x2x2 and $p_{max}=9$ for 3x3x3\footnote{\REV{We limit ourselves to up to 9 twists here, because our network has not enough capacity to learn all states of 3x3x3 Rubik’s Cube. Experiments with higher twist numbers \REVB{during training} did not improve the solved-rates.}}\REV{, both in half-turn metric}.  This covers the complete cube space for 2x2x2, but only a small subset for 3x3x3, where God's number~\cite{rokicki2014diameter} is known to be 20. We evaluate the trained agents on 200 scrambled cubes that are created by applying a given number $p$ of scrambling twists to a solved cube. The agent now tries to solve each scrambled cube. A cube is said to be \textit{unsolved} if the agent cannot reach the solved cube in $e_E=20$ steps. More details on our method are found in~\cite{konen22b}.

Here we are interested in the relative strength of agents with and without MCTS wrapping. The results are shown in  Fig.~\ref{fig:Rubiks-2x2-3x3}: While \TCLbase could only solve 75\% (2x2x2) or 25\% (3x3x3) of the scrambled cubes, resp., the MCTS-wrapped agent \TCLwrap could either fully solve the problem (2x2x2) or at least double or triple the percentage of solved cubes (3x3x3). 

\textit{Interpretation 2x2x2}: Since the solved-rate of \TCLbase is only 75\%, the value function $V(s)$ does not predict the right action for every state $s$ (resulting in a short path to the solved cube). However, if we add the planning stage of MCTS-wrapper, then the action with the highest $V(s)$ after a few `what-if' steps is selected. This is sufficient to boost the solved-rate to 100\% after 200 or more MCTS-iterations.

\textit{Interpretation 3x3x3}: The agent has seen during training only a small subset of cubes with up to 9 scrambling twists. Therefore, the solved-rates for $p \in {8,9}$ are much lower for \TCLbase  because it is very likely that the cube `escapes' with a wrong move into the unknown area of $p=10$ or higher. It is interesting to see that the MCTS planning stage can double or triple the solved-rate. However, it can not cure everything since the high branching factor of 18 together with slight inaccuracies of the value function approximator makes it likely that even 1000 iterations of MCTS-wrapper do not explore enough to find the right path.

\begin{table}[tbp]
    \caption{Training times to train $n_{ag}$ agents without and with MCTS wrapping (hypothetical) for all games.
    } 
    \label{tab:compTimes}
    \centering
    \begin{footnotesize}
    \begin{tabular}{|l|r|r|r|r||r|} \hline\hline
    game & $n_{ag}$ & 
    \begin{minipage}{1.0cm}{base${}^{\vphantom{R}}$ training time${}_{\vphantom{M}}$} \end{minipage}
        & $i_{MCTS}$ & \textit{factor}
        & \begin{minipage}{1.5cm}{ hypothetical${}^{\vphantom{R}}$ wrapped training time${}_{\vphantom{M}}$} \end{minipage}  \\ \hline
Othello    & 20 & 1.5 d& 10,000    & 2,575 & 10.6 years \\ \hline
ConnectFour& 10 & 1.4 d&  1,000    &   850 &  3.3 years \\ \hline
RubiksCube &  5 & 2.2 h&  1,000    &   770 &   71 days \\ \hline\hline
    \end{tabular}
    \end{footnotesize}
\end{table}

\subsection{Computation times}
\label{sec:comp_times}

The MCTS wrapper for RL agents, as proposed in this paper, has the advantage that it does not cost any additional training time since it is an enhancement added \textit{after} agent training.

The extra computational resources needed during game play or evaluation are moderate. This is because \REV{we usually only need} a few evaluation episodes (compared to the huge number of training episodes) and \REV{for these few episodes the 10,000 iterations are not a large computational burden}\REVB{, as shown in Tab.~\ref{tab:computation-time-per-move} (note the unit milliseconds)}.

\begin{table*}[htb]
    \caption{
    \REVB{Move times of \TCLbase, \TCLwrap, and \textit{MCTS}, where 
    \TCLwrap[$i$] and \textit{MCTS} [$i$] refer to the respective agents
    using  $i$ 
    iterations. 
    Shown are average times and standard deviations in milliseconds. For the games Othello and ConnectFour, the values were averaged over 50 games. 
    For RubiksCube, the values are based on 25 cubes, each initially scrambled with 8 twists.
    }}
    \label{tab:computation-time-per-move}
    \centering
    \begin{footnotesize}
    \begin{tabular}{|l|c|c|c|c|c|} \hline\hline
    game & \TCLbase & \TCLwrap[100] & \TCLwrap[1000] & \TCLwrap[10000] & \textit{MCTS} [10000] \\ \hline\hline
    Othello & $0.4\pm 0.5$  & $10\pm 6$ & $100 \pm 80$ & 1,300 $\pm$ 700 & 551 $\pm$ 371\\
    
    ConnectFour & $0.05\pm0.02$  & $3\pm 2$ & $30 \pm 20$ & $300 \pm 200$ & 66 $\pm$ 20 \\

    3x3x3 Rubik's & $0.6\pm0.7$  & $10 \pm 6$  & $100 \pm 70$  & 1,800 $\pm$ 1,000 & 783 $\pm$ 49 \\
    \hline\hline
    \end{tabular}
    \end{footnotesize}
\end{table*}

The above advantage becomes more apparent if we compare the actual \TCLbase training times with the would-be training times if the MCTS planning stage were also used during training, as shown in Tab.~\ref{tab:compTimes}: The base training time is the time actually needed to train $n_{ag}$ agents without MCTS wrapper. All computations were done on a single CPU Intel i7-9850H @ 2.60GHz.\footnote{To get a single-agent training time, the base training time has to be divided by $n_{ag}$ which results for example in 1.8 hours training time for one Othello agent.} 

The wrapped training times are estimated by multiplying the base training time with \textit{factor} which is established by running a few episodes without and with MCTS wrapper doing $i_{MCTS}$ iterations. \REV{This estimate assumes that a wrapped agent needs as many training episodes as a base agent. This assumption needs not to be true, a wrapped agent could reach similar performance in fewer episodes. We  experimentally investigate and discuss this point further in Sec.~\ref{sec:mcts_train}. If this assumption were true, the hypothetical training times would be astronomical:}
We see from Tab.~\ref{tab:compTimes} that with the same hardware, many years or at least near 100 days of computation time would be necessary. Of course, large speed-ups are possible with dedicated hardware or parallel execution on many cores, but often this hardware is just not available.   

\subsection{\REV{MCTS Inside Self-Play Training}}
\label{sec:mcts_train}
\REV{MCTS inside self‐play training can improve the quality of the experience generated. It might be that substantially fewer episodes are necessary to learn the same (or a better) function.}

\REV{In order to investigate this, we conducted several experiments for the Othello case: We cannot afford 10,000 MCTS iterations in the training loop for 250,000 episodes, as Tab.~\ref{tab:compTimes} shows, but we can approach it from several directions.}

\begin{figure}[htb]%
\centerline{
\includegraphics[width=1.0\columnwidth]{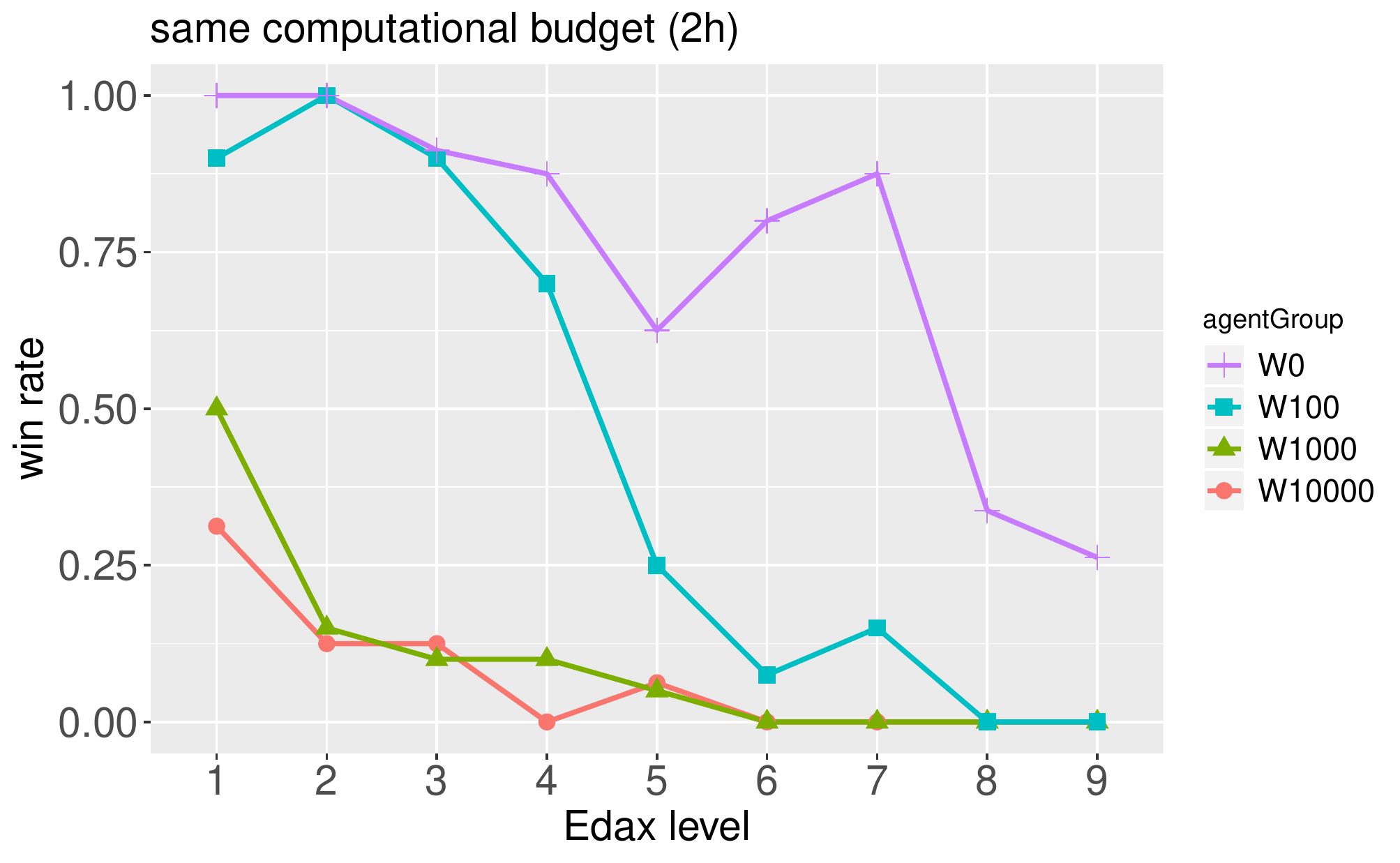}}
\caption{
\REV{MCTS inside training under the same training time budget (2h) as \TCLbase. The label W$*$ stands for MCTS with $*$ = 0, 100, 1,000, 10,000 iterations inside self-play training. This allows for 250,000, 12,000, 1,200 and 120 episodes, resp. W0 is identical to \TCLwrap.
Shown are the win rates vs. Edax from 20 competition runs (40 for W0) conducted in the same way as in Fig.~\ref{fig:MCTSWrap-TCL}. Each agent uses 10,000 iterations during Edax evaluation.
} }
\label{fig:MCTS-W100}%
\end{figure}

\begin{figure}[htb]%
\centerline{
\includegraphics[width=1.0\columnwidth]{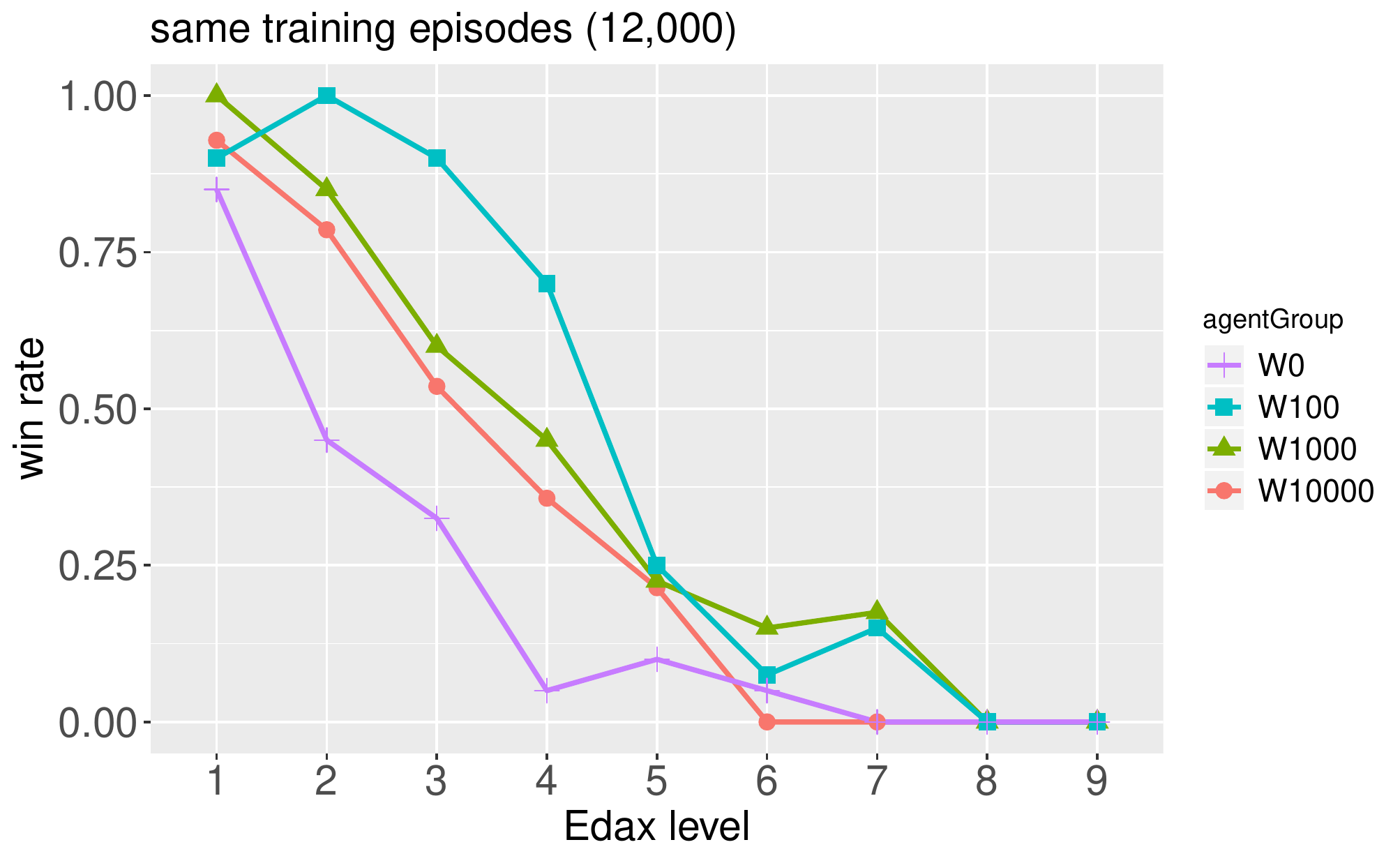}
}
\caption{
\REV{MCTS inside training with the same small number of training episodes (12,000). The labels W$*$ have the same meaning as in Fig.~\ref{fig:MCTS-W100}. Note that W0 is now different from \TCLwrap (only 12,000 training episodes). Shown are the win rates from 20 competition runs (16 in the case of W10000).
} }
\label{fig:MCTS-W100-12k}%
\end{figure}

\begin{figure}[htb]%
\centerline{
\includegraphics[width=1.0\columnwidth]{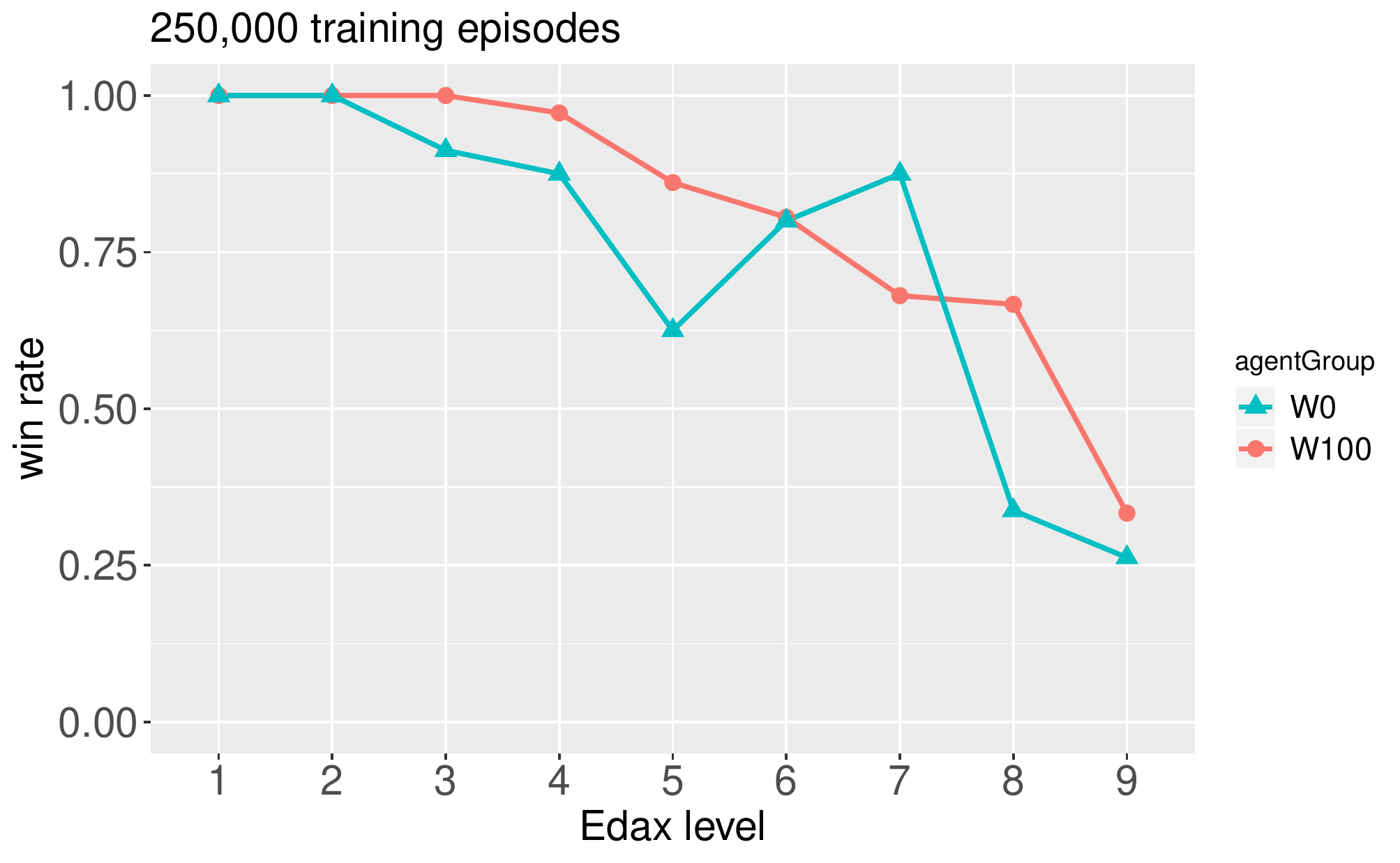}
}
\caption{
\REV{MCTS inside training with the same number of training episodes as \TCLwrap (250,000). The labels W$*$ have the same meaning as in Fig.~\ref{fig:MCTS-W100}. W0 is identical to \TCLwrap. Shown are the win rates vs. Edax from 40 competition runs.
} }
\label{fig:MCTS-W100-250k}%
\end{figure}

\REV{First, we present in Fig.~\ref{fig:MCTS-W100} several runs with the same computational training budget (2 hours) as \TCLbase. All agents, regardless of their MCTS iterations during training, use the \textit{same} number of 10,000 MCTS iterations during Edax evaluation (the standard for \TCLwrap in Othello). It is not surprising that W1000 and W10000 cannot compete with Edax  on any level, since the number of training episodes is too small to explore a significant portion of the state space.}

\REV{The next experiment requires a higher computational budget: We take for all agents the same number of 12,000 training episodes and allow for different numbers
of MCTS iterations during training (Fig.~\ref{fig:MCTS-W100-12k}). This leads to single-agent training times of 1.2 days 
for W1000 and \textit{11.5 days} 
for W10000,
which is 16$\times$ and 153$\times$ the training time of \TCLwrap, resp.
Nevertheless, the win rates against Edax  level 5--9 are considerably lower than for \TCLwrap. Surprisingly, the win rates for W1000 and W10000 are for some levels lower than for W100. 
 \REVB{It might be that the agents with higher iteration counts become too deterministic and explore too little during self-play.}\footnote{\REVB{We used $\epsilon$-greedy exploration during self-play and tuned $\epsilon$ on the W100 agent, where we found a very small $\epsilon$ (compatible with 0) to give the best results.}}
}

\REV{The third experiment tests MCTS iterations during training when we choose the same number of training episodes (250,000) as \TCLwrap. 100 iterations are the maximum possible here, since the training time for a single agent is already 2.5 days.  
The results in Fig.~\ref{fig:MCTS-W100-250k} show that W100 is mostly better than W0 (\TCLwrap), with the largest increase for Edax level 8 (from 34\% to 66\%). 
} 

\REV{In summary, these experiments show that MCTS inside self-play training cannot reach the same results as \TCLwrap under (a)~the same computational budget (2 hours) or (b)~a small number (12,000) of training episodes. 
But if we use the same number of 250,000 training episodes as \TCLwrap, we see a small positive effect of using MCTS in the training loop. 
However, this requires 30 times more computing time.}
\section{Discussion}
\label{sec:discuss}

\subsection{Related work}
\REV{Self-play RL has a long tradition in game learning, with Tesauro's TD-Gammon~\cite{tesauro1994td} being a very early TD-learning application to Backgammon.}
The seminal papers of Silver et al. on 
AlphaGo and AlphaZero  \cite{silver2016AlphaGo,silver2017AlpaZeroChess} \REV{lifted this for the games Go, chess and shogi to a new complexity and \REVB{performance} level. They} have stirred the interest of many researchers to achieve similar things with smaller hardware requirements and/or fewer training cycles. Thakoor et al.~\cite{ThakoorNair2017learning} provided in 2017  a general AlphaZero implementation in Python with less computational demands than the original. 
But even their architecture requires \REV{for 6x6 Othello} 3 days of training on a specialized cloud computing service (Google Compute Engine with GPU support). 
Several works of Wang et al. \cite{wang2019alternative,wang2019hyper,wang2020warm} focus on different aspects of the AlphaZero architecture: alternative loss functions, hyperparameter tuning and warm-start enhancements. They test these aspects
on smaller games like 6x6 Othello or 5x5 ConnectFour. The work of Chang et al.~\cite{chang2018-BigWin} covered several AlphaZero improvements applied to 6x6 Othello. \REV{van der Ree and Wiering~\cite{Ree2013reinforcement} investigate TD-, Q- and Sarsa-learning for 8x8 Othello with a simple neural network (one hidden layer with 50 neurons).}

Dawson~\cite{dawson2020} introduces a CNN-based and AlphaZero-inspired RL agent named ConnectZero for ConnectFour, which can be played online and which reaches a good playing strength against MCTS$_{1000}$. Young et al.~\cite{young2018Lessons} report on an Alpha\-Zero implementation applied to ConnectFour. Here, training took between 21 and 77 hours of GPU time.

Recently in 2022, 
Norelli and Panconesi~\cite{norelli2022olivaw} \REV{presented an approach} that is close to our work: They as well pursue the goal to set up an AlphaZero-inspired algorithm at much lower cost than the original AlphaZero~\cite{silver2017AlpaZeroChess}. The agent in~\cite{norelli2022olivaw} \REV{is based on a residual DNN} and is trained solely by self-play. It is able to play 8x8 Othello and to defeat the strong Othello program Edax~\cite{delorme2019} up to level 10. Although much less computationally demanding than the original AlphaZero~\cite{silver2017AlpaZeroChess}, their training time took roughly one month on Colaboratory, a free Google cloud computing service offering GPUs and TPUs. \REV{See  Sec.~\ref{sec:comparisonRL} for a direct comparison between Norelli and Panconesi~\cite{norelli2022olivaw} and our work.}

Apart from Norelli and Panconesi~\cite{norelli2022olivaw}, there are only few works on Othello game learning that actually benchmark against Edax: Liskowski et al.~\cite{liskowski2018learning} presented in 2018 an agent obtained by training a convolutional neural network (CNN) with the help of a database of expert moves. Their agent could defeat Edax up to and including level 2. 

Our work presented here is based on an earlier Bachelor thesis~\cite{Scheier2020b} published in 2020 (but only in German); it presents an n-tuple RL agent for Othello trained in 1.8 hours on standard hardware (no GPU) that can defeat Edax up to and including level 7.  

\REV{For the puzzle Rubik's Cube, the pioneering work of McAleer~\cite{mcaleer2018solving} and Agostinelli~\cite{agostinelli2019solving} in 2018 and 2019 shows} that the 3x3x3 cube can 
be solved without putting human knowledge or positional-pattern databases into the agent. They solve arbitrary scrambled cubes with a method that is partly inspired by AlphaZero but also contains special tricks for Rubik's Cube. 

A work related to GBG~\cite{Konen2019b,Konen22a} is the general game system Ludii~\cite{Piette2019}. 
Ludii is an efficient general game system based on a ludeme library implemented in Java, allowing to play as well as to generate a large variety of strategy games. Currently, all AI agents implemented in Ludii are tree-based agents (MCTS variants or AlphaBeta). GBG, on the other hand,
offers the possibility to train RL-based algorithms on several games.

\REV{Soemers et al.~\cite{soemers2021deep} describe a bridge between 
Ludii~\cite{Piette2019} and Polygames~\cite{cazenave2020polygames}, the latter providing DNN algorithms for strategy games. 
Similar to our work, they couple approximator networks (DNNs) with MCTS, but for different games. 
With 20 hours of training time, 8 GPUs, 80 CPU cores, and 475 GB of memory allocation per training job, their resource usage is in a different dimension than our training process. 
}


\subsection{\REV{Related Work in} N-Tuple Research}
\label{sec:comparisonNTuple}
N-tuple networks, which are an important building block of our approach, have shown to work well in many games, e.~g., in ConnectFour~\cite{Bagh15,Thil14}, O\-thel\-lo~\cite{Lucas08}, EinStein würfelt nicht (EWN)~\cite{Chu2017EinStein}, 2048~\cite{szubert2014temporal}, SZ-Tetris~\cite{jaskowski2015high}, etc.
Other function approximation networks (DNN or other) could be used as well in AlphaZero-inspired RL, but n-tuple networks have the advantage that they can be trained very fast on off-the-shelf hardware.

There are two papers in the game learning literature that combine n-tuple networks with MCTS:
Sironi et al.~\cite{sironi2018self} use the n-tuple bandit EA to automatically tune a self-adaptive MCTS. This is an interesting approach but for a completely  different goal and not related to AlphaZero. 

Chu et al.~\cite{Chu2017EinStein} use an n-tuple network as a \REV{guidance 
for MCTS in the game EWN. They train an n-tuple network via Monte-Carlo RL and incorporate the trained heuristic with three different approaches into MCTS. Monte Carlo is a special form of temporal difference (TD) learning, namely TD(1). To the best of our knowledge, our work is the first to couple n-tuple networks with MCTS using TD($\lambda$)-training with arbitrary $\lambda$.}


\REV{Neither Chu et al.~\cite{Chu2017EinStein} nor we use MCTS in the training loop for the main results. However, we compare in Sec.~\ref{sec:mcts_train} -- at least for some viable settings --  MCTS within and outside of self-play training. Finally, we use our approach for other games than those studied in~\cite{Chu2017EinStein}. }

\subsection{Comparison with Other RL Research}
\label{sec:comparisonRL}

In this section, we compare our results with other RL approaches from the literature.

\REV{Concerning the game ConnectFour, it was shown in Sec.~\ref{sec:result-C4} that Dawson's ConnectZero lost most or all of its episodes when starting against AlphaBeta-DL or \TCLwrap.
} 
This is in contrast to our \TCLbase and \TCLwrap, which win nearly all episodes when starting against AlphaBeta-DL (see Tab.~\ref{tab:C4-4agent-tourn}). 

Concerning the game Othello, there are a number of other researchers that do RL-based game learning: 
van der Ree and Wiering~\cite{Ree2013reinforcement} reached in 2013 with their Q-learning agent against the heuristic player \textsc{Bench} (positional player) a win rate of 87\%. We reach with both \TCLbase and \TCLwrap a win rate of 100\% against \textsc{Bench}. 
Liskowski et al.~\cite{liskowski2018learning} show in Table IX that their agent wins against Edax up to and including Edax level 2. We win up to and including Edax level~7. 

In 2022,
Norelli and Panconesi~\cite{norelli2022olivaw} obtained with their system \textsc{Olivaw} the best Othello results up-to-date: It defeats Edax up to and including Edax level 10. This is a truly impressive result, but it also took considerable computational resources to achieve it: Although much cheaper than DeepMind's original AlphaZero, they needed an informal crowd computing project with 19 people for game generation and then about 30 days to train a single agent on Google Colaboratory using GPU and TPU hardware \REV{(50,000 training episodes with 100-400 MCTS iterations). Due to the large training time}, fine-tuning of hyperparameters or ablation studies could not be undertaken.

In our work presented here, we defeat Edax only up to level~7, but with a much simpler architecture that is trainable in less than 2 hours on a single standard CPU.\footnote{\REV{level 8 with MCTS in the training loop (Fig.~\ref{fig:MCTS-W100-250k}) and 60 hours training}} It is, on the one hand, interesting that our architecture, which keeps the costly MCTS completely out of the training process, can get so far. 

On the other hand, there is of course a performance gap to~\cite{norelli2022olivaw}, and it would be interesting to investigate which element of the more complex architecture in ~\cite{norelli2022olivaw} is responsible for the performance gain. We see here two hypothetical candidates: First, including MCTS in the training phase leads to better positional material in the replay buffer. Second, the network architecture of \textsc{Olivaw} uses a Residual Network, a somewhat reduced version of the original AlphaZero Residual Network, but still a deeper architecture than our n-tuple network.

\REV{Concerning the puzzle Rubik's Cube,  the deep network used by McAleer~\cite{mcaleer2018solving} and Agostinelli~\cite{agostinelli2019solving}} had over 12 million weights and was trained for 44 hours on a 32-core server with 3 GPUs. Our approach \REV{with much less computational effort} can solve the 2x2x2 cube completely, but the 3x3x3 cube only partly.

%

\section{Conclusion and Future Work}
\label{sec:conclusion}

We have shown on the three challenging games, Othello, ConnectFour, and Rubik's Cube, that an AlphaZero-inspired MCTS planning stage boosts the performance of TD-n-tuple networks. Interestingly, this performance boost is even reached when MCTS is \textit{not} part of the training stage, which leads to very large reductions in training times and computational resources.  

The new architecture was evaluated on the three games without any game-specific changes. We reach \REVB{near-}perfect play for ConnectFour and 2x2x2 Rubik's Cube. For the games Othello and 3x3x3 Rubik's Cube, we observe good results and increased performance compared to our version without 
MCTS planning stage, but we do not reach the high-quality results of Norelli and Panconesi~\cite{norelli2022olivaw} on Othello (beats Edax level 10 where we reach only level 7) and of Agostinelli, McAleer et al.~\cite{mcaleer2018solving,agostinelli2019solving} on 3x3x3 Rubik's Cube (they solve all scrambled cubes while we solve only cubes with up to 9 twists). Both high-performing approaches require considerably more computational resources. 

\REV{We compared the effects of using MCTS within or outside of self-play training. For Othello and n-tuple networks, we have found that large increases in computation time are accompanied by no or only small increases in performance.}

It is an interesting topic of future research to investigate which element of the more complex architecture (MCTS in the training phase or deep residual network for the approximator) is more relevant to reach the impressive high-quality results of others. 
However, for moderately complex games, our smaller architecture has the advantage of allowing faster training and more parameter tuning with simpler hardware that is accessible to anyone.

We also plan to extend our MCTS wrapper concept  to non-deterministic games (e.~g., EWN, 2048, Blackjack, Poker) where previous research~\cite{Kutsch17} has shown that plain MCTS is not sufficient and has to be extended by the Expectimax approach.

\section*{Acknowledgment}
\REVB{The authors would like to thank the anonymous reviewers for their insightful comments that helped a lot to shape and improve this work.}



%
%
%


\end{document}

%% file: rl-with-mcts.tex
\tikzstyle{block} = [draw, fill=blue!20, rectangle, 
    minimum height=3em, minimum width=6em]
\tikzstyle{input} = [coordinate]
\tikzstyle{arrow} = [->,very thick,>=stealth]

\begin{center}
\begin{tikzpicture}[auto, node distance=2cm and 1cm]
    \node [input] (action) {};															
		\node [above =0cm of action] (ta) {action $a_t$}; 			
    \node [block, right of=action,yshift=-1cm] (env) {environment};
		\node [input, left of=env, yshift=-1.2cm] (observ) {};	
		\node [above =0cm of observ] (to) {observation $s_t$}; 
		\node [below =0cm of observ] (tr) {reward $r_t$}; 
    \node [block, left of=observ,yshift=+1.2cm] (agent) 		
			{\begin{tabular}{l}
			 MCTS wrapper \\[0.2cm]
			 \hspace{0.5cm}\begin{tikzpicture}
			    \node [minimum height=1em, draw] {RL agent};
			 \end{tikzpicture}
			 \end{tabular}
			};

    \draw [arrow,->] (action) -|  (env);			
    \draw [arrow,->] (env) |-  (observ);
    \draw [arrow,->] (observ) -| (agent);
    \draw [arrow,->] (agent) |- (action);
\end{tikzpicture}
\end{center}